\documentclass[twoside,leqno,twocolumn]{article}  
\usepackage{ltexpprt} 

\usepackage[noadjust]{cite}
\usepackage{bbm}
\usepackage{hyperref}
\usepackage{caption}
\usepackage{color}
\usepackage{xspace}
\usepackage[utf8x]{inputenc}
\usepackage{graphicx}
\usepackage[cmex10]{amsmath}
\usepackage{amssymb}
\usepackage{booktabs}
\usepackage{multirow}
\usepackage{subfig}
\usepackage{verbatim}
\usepackage{wrapfig}

\usepackage{booktabs}
\usepackage{microtype}

\newcommand*\justify{%
	\fontdimen2\font=0.4em
	\fontdimen3\font=0.2em
	\fontdimen4\font=0.1em
	\fontdimen7\font=0.1em
	\hyphenchar\font=`\-
}

\begin{document}
%
\title{Exploring Latent Semantic Factors to Find Useful Product Reviews}

\newcommand{\squishlist}{
   \begin{list}{$\bullet$}
    { \setlength{\itemsep}{0pt}      \setlength{\parsep}{3pt}
      \setlength{\topsep}{3pt}       \setlength{\partopsep}{0pt}
      \setlength{\leftmargin}{1.5em} \setlength{\labelwidth}{1em}
      \setlength{\labelsep}{0.5em} } }
\newcommand{\squishlisttwo}{
   \begin{list}{$\bullet$}
    { \setlength{\itemsep}{0pt}    \setlength{\parsep}{1pt}
      \setlength{\topsep}{1pt}     \setlength{\partopsep}{0pt}
      \setlength{\leftmargin}{1em} \setlength{\labelwidth}{0.5em}
      \setlength{\labelsep}{0.5em} } }

\newcommand{\squishend}{
    \end{list} 
}




%

\author{Subhabrata Mukherjee\footnotemark[1] \and Kashyap Popat\footnotemark[1] \and Gerhard Weikum\thanks{Max Planck Institute for Informatics \newline Saarland Informatics Campus, Germany \newline Email: \{smukherjee, kpopat, weikum\}@mpi-inf.mpg.de
}
}

\maketitle

\begin{abstract}
\small\baselineskip=9pt
Online reviews provided by consumers are a valuable asset for e-Commerce platforms, influencing potential consumers in making purchasing decisions. However, these reviews are of varying quality, with the {\em useful} ones buried deep within a heap of non-informative reviews. In this work, we attempt to automatically identify review quality in terms of its {\em helpfulness} to the end consumers. In contrast to previous works in this domain exploiting a variety of syntactic and community-level features, we delve deep into the {\em semantics} of reviews as to what makes them useful, providing {\em interpretable} explanation for the same. We identify a set of {\em consistency} and {\em semantic} factors, all from the {\em text, ratings, and timestamps} of user-generated reviews, making our approach generalizable across all communities and domains. We explore review semantics in terms of several latent factors like the {\em expertise} of its author, his judgment about the fine-grained {\em facets} of the 
underlying 
product, and his {\em writing style}. These are cast into a Hidden Markov Model -- Latent Dirichlet Allocation (HMM-LDA) based model to {\em jointly} infer: (i) reviewer expertise, (ii) item facets, and (iii) review helpfulness. Large-scale experiments on {\em five} real-world datasets from \textit{Amazon} show significant improvement over state-of-the-art baselines in predicting and ranking useful reviews. 

\end{abstract}

\section{Introduction}

\noindent {\bf Motivation:} With the rapid growth in e-Commerce, product reviews have become a crucial component for the business. As consumers cannot test the functionality of a product prior to purchase, these reviews help them make an informed decision to buy the product or not. As per a survey conducted by Nielsen Corporations, $40\%$ of online consumers indicated that they would not buy electronics without consulting online reviews first~\cite{nielsen}. 

Due to the increasing dependency on user-generated reviews, it is crucial to understand their quality --- that can widely vary from being an excellent-detailed opinion to superficial criticizing or praising, to spams in the worst case. Without any indication of the review quality, it is overwhelming for consumers to browse through a multitude of reviews. In order to help consumers in finding useful reviews, most of the e-Commerce platforms nowadays allow users to vote whether a product review is helpful or not. For instance, any {\em Amazon} product review is accompanied with information like $x$ out of $y$ users found the review helpful. {\em This {helpfulness score} ($x/y$) can be considered as a proxy for the review quality, and its {usefulness} to the end customers.} In this work, we aim to automatically find the helpfulness score of a review based on certain consistency, and semantic aspects of the review like: whether the review is written by an expert, what are the important facets of the product 
outlined in his review, what do other 
experts have to say about the given product, timeliness of the review etc. --- that are automatically mined as latent factors from review texts.

\section{Related Research and their Limitations}
\noindent{\bf Predicting Review Helpfulness and Spams}: Prior works on predicting review usefulness mostly operate on shallow syntactic textual features like bag-of-words, part-of-speech tags, and tf-idf (term, and inverse document frequency) statistics~\cite{Kim:2006:AAR:1610075.1610135,Lu:2010:ESC:1772690.1772761}. These works, and other related works on finding review spams~\cite{Liu2008, Liu2013} classify extremely opinionated reviews as not helpful. Similarly, other works exploiting rating \& activity features like frequency of user posts, average ratings of users and items~\cite{O'Mahony:2009:LRH:1639714.1639774,Lu:2010:ESC:1772690.1772761, liu-EtAl:2007:EMNLP-CoNLL2007} consider extreme ratings and deviations as indicative of unhelpful reviews. 
Some recent works incorporate additional information like community-specific characteristics (who-voted-whom) with explicit user network~\cite{Tang:2013:CRH:2507157.2507183,Lu:2010:ESC:1772690.1772761}, and item-specific meta-data like {\em explicit} item facets 
and product brands~\cite{icdm2008,Kim:2006:AAR:1610075.1610135}. 

Apart from the requirement of a large number of meta-features that restrict the generalizability of many of these models to any arbitrary domain, these shallow features do not 
analyze what the review is {\em about}, and, therefore, cannot {\em explain} why it should be helpful for a given product. Some of these works~\cite{O'Mahony:2009:LRH:1639714.1639774,icdm2008} identify {\em expertise} of a review's author as an important feature. However, in absence of suitable modeling techniques, they consider prior reputation features like user activity, and low rating deviation as proxy for user expertise. 

\noindent{\bf Latent Factors for Review Analysis}: Prior approaches for analyzing review texts aim to learn latent topics~\cite{linCIKM2009}, latent aspects and their ratings~\cite{lakkarajuSDM2011,wang2011,mcauleyrecsys2013}, and user-user interactions~\cite{West-etal:2014}. The author writing style is also used in \cite{mukherjeeSDM2014}. However, these prior approaches do not factor in the temporal dynamics and user expertise.

\noindent{\bf Modeling Expertise}: Our model for capturing user expertise draws motivation from~\cite{mukherjee2015jertm, mukherjee2016KDD, mcauleyWWW2013} with significant differences. \cite{mcauleyWWW2013} ignores the content of reviews, and focuses only on the {\em rating behavior} for modeling expertise evolution, which is addressed in \cite{mukherjee2015jertm}. \cite{mukherjee2016KDD} proposes a generalized (continuous analog) version of expertise evolution model over~\cite{mukherjee2015jertm}. However, it is much more complex and computationally expensive. Therefore, in this work, we use the simpler version~\cite{mukherjee2015jertm} addressing some of its computational deficiencies (refer to Section~\ref{subsec:diff} for details) for tractable inference in our problem setting.  


\section{Overview of our Approach and Contributions}



Our work aims to overcome the limitations of prior works by exploring the {\em semantics} and {\em consistency} of a review to predict its {\em helpfulness} for a given product. The first step towards understanding the {\em semantics} of a review is to uncover the facet descriptions of the target product outlined in the review. We treat these facets as {\em latent} and use Latent Dirichlet Allocation (LDA) to discover them as topic clusters. The second step is to find the {\em expertise} of the users who wrote the review, and their description of the different (latent) facets of the product. 
In this work, we model expertise as a {\em latent} variable that evolves over time using Hidden Markov Model (HMM). 

We make use of {\em distributional hypotheses} like: expert users agree on what are the important facets of a product, and their description (or, writing style) of those facets 
influences the helpfulness of a review. We also derive several {\em consistency} 
features --- all from the given quintuple $\langle$\textit{userId, itemId, rating, reviewText, timepoint}$\rangle$ --- like prior user reputation, item prominence, and timeliness of a review. 
Finally, we leverage the interplay between all of the above factors in a {\em joint} setting to predict the review helpfulness.

For interpretable explanation, we derive interesting insights from the latent word clusters used by experts --- for instance, reviews describing the underlying ``theme and storytelling'' of {\em movies} and {\em books}, the ``style'' of {\em music}, and ``hygiene'' of {\em food} are considered most helpful for the respective domains.

\noindent In summary, we make the following novel contributions:
a) {\bf Model:} We propose an approach to leverage the {\em semantics} and {\em consistency} of reviews to predict their helpfulness. We propose a Hidden Markov Model -- Latent Dirichlet Allocation (HMM-LDA) based model that jointly learns the (latent) product facets, (latent) user expertise, and his writing style from {\em observed} words in reviews at explicit timepoints.\\
b) {\bf Algorithm:} We introduce an effective learning algorithm based on an iterative stochastic optimization process that reduces the mean squared error of the predicted helpfulness scores with the ground scores, as well as maximizes the log-likelihood of the data.\\
c) {\bf Experiments:} We perform large-scale experiments with real-world datasets from {\em five} different domains in {\em Amazon}, together comprising of $29$ million reviews from $5.7$ million users on $1.9$ million items, and demonstrate substantial improvement over state-of-the art baselines for {\em prediction} and {\em ranking} tasks.




\section{Review Helpfulness Factors}\label{sec:features}
In this section, we outline the components of our model that analyze the {\em semantics} and {\em consistency} of reviews. 

\noindent {\bf Item Facets: }
It is essential to understand the different facets of an item in a review. For instance, a camera review can focus on facets like ``resolution", ``zoom", ``price", ``size'', or a movie review can focus on ``narration", ``cinematography", ``acting", ``direction'' etc. However, not all facets are equally important for an item. For example, a review downrating a camera for ``late delivery'' by the seller is not as helpful to end consumers as opposed to downrating it due to {\em grainy resolution} or {\em shaky zoom}. Therefore, a helpful review should focus on the {\em important facets} of an item. 
We model facets as {\em latent} variables, 
where the item's latent facet distribution in the review is indicative of how {\em detailed} and {\em diverse} the review is.

\noindent {\bf Review Writing Style: }
{\em Words} used to describe the facets play a crucial role in making the review useful to the consumers. 
An important aspect of an expert writing style is to use {\em precise, domain-specific} vocabulary to describe a facet in {\em details}, rather than using generic words. For instance, contrast this {\em expert} camera review: ``{\tt \small \justify{60D focus screen is `grainy'. It is the `precision matte' surface that helps to increase contrast and minimize depth of field for manual focusing. The Ef-s screen is even more so for use with fast primes...}}''
with this {\em amateur} one: ``{\tt \small \justify{This camera is pure garbage. It is the worst 
one I have ever owned. I bought it last xmas on a deal and have thrown it away and replaced it with a decent camera.}}"
We learn a language model from the latent facets and user expertise that helps to distinguish the writing style of an experienced user from an amateur one. 



\noindent {\bf Reviewer Expertise: }\label{subsec:revExp}
Prior works~\cite{Liu2008,O'Mahony:2009:LRH:1639714.1639774} used proxy features like user activity and reputation (e.g. number of reviews written, feedback from community etc.) to harness users' expertise under the hypothesis that expert reviews are positively correlated to review helpfulness. 
We {\em explicitly} model user expertise, drawing motivation from recent works~\cite{mcauleyWWW2013, mukherjee2015jertm, mukherjee2016KDD} with substantial modifications for tractable inference (refer Section~\ref{subsec:diff}). 
Expertise is not static, but evolves over time. 
A user who was amateur at the time of entering a community, may have become an expert now.
We model {\em expertise} as a {\em latent} variable that evolves over time, exploiting the hypothesis that users at similar levels of expertise have similar rating behavior, facet preferences, and writing style. 

\noindent {\bf Distributional Hypotheses: } 
\label{subsec:distrhyp}
Our approach makes use of the following hypotheses to capture helpfulness:\\
\noindent i) Reviews (e.g., camera reviews) with similar facet distribution (e.g., focusing on ``zoom'' and ``resolution'') for items are likely to be equally helpful. 
\\
\noindent ii) Users with similar facet preferences and expertise are likely to be equally helpful. 

In traditional collaborative filtering approaches for recommender systems, (i) and (ii) are similar to item-item and user-user similarities, respectively.

\noindent {\bf Consistency: }
Users and items do not gain reputation overnight. 
Therefore prior reputation of users and items are good indicators of associated reviews' helpfulness. We use the following features to guide our model in learning {\em latent} distributions based on review helpfulness.

{\noindent \em Prior user reputation:} Average helpfulness votes received by the user's past reviews from other users.\\ 
{\noindent \em Prior item prominence:} Average helpfulness votes received by the item's past reviews from other users, which is also indicative of the {\em prominence} of the item. \\
{\noindent \em User rating deviation:} Absolute deviation between the user's rating on an item, and average rating assigned by the user over all other items. This captures the mean user rating behavior, and, therefore, scenarios where the user is too dis-satisfied (or, otherwise) with an item.\\
{\noindent \em Item rating deviation:} Absolute deviation between a user's rating on the item, and average rating received by the item from all other users. This captures the scenario where a user unnecessarily criticizes or praises the item.\\
{\noindent \em Global rating deviation:} Absolute deviation between a user's rating on an item, and average rating of all items by all users in the community --- capturing deviation of user behavior from general community behavior.\\
\noindent {\em Timeliness or ``Early-bird'' bias: }
Prior work~\cite{icdm2008} shows a positive influence of a review's publication date on its perceived helpfulness. Early and ``timely'' reviews are more useful when the item is launched for consumers to make an informed decision. Also, early reviews are exposed to consumers for a longer period of time allowing them to garner more votes over time, compared to recent reviews. The timestamp of the {\em first} review on a given item $i$ is considered to be the reference timepoint (say, $t_{i,0}$). Therefore, the timeliness of any other review on the item at time $t_i$ is computed as: $exp^{-(t_i - t_{i,0})}$.

The following section depicts our approach to model all of these factors {\em jointly} to 
predict review helpfulness.

\section{Joint Model for Review Helpfulness}\label{sec:model}

\subsection{Incorporating Consistency Factors} \label{subsec:consist}
Let $u \in U$ be a user writing a review at time $t \in T$ on an item $i \in I$. Let $d=\{w_1, w_2,... w_{|N_d|}\}$ be the corresponding review text with a sequence of words $\langle w \rangle$, and rating $r \in R$. Each such review is associated with a helpfulness score $h \in [0-1]$. Let $b_t$ be the corresponding timeliness of the review computed as  $exp^{-(t - t_{i,0})}$, where $t_{i,0}$ is the {\em first} review on the item $i$.

Let $\beta_u$ be the average helpfulness score of user $u$ over all the reviews written by her (capturing user reputation), and $\beta_i$ be the average helpfulness score of all reviews for item $i$ (capturing item prominence). Let $\overline{r}_u$ be the average rating assigned by the user over all items, $\overline{r}_i$ be the average rating assigned to the item by all users, and $\overline{r}_g$ be the average global rating over all items and users. {\em Consistency} features include prior item and user reputation, deviation features, and burst. 

Let $\xi$ be a tensor of dimension $E \times Z$, where $E$ is the number of expertise levels of the users, and $Z$ is the number of latent facets of the items. $\xi_{e,z}$ depicts the opinion of users at (latent) expertise level $e \in E$ about the (latent) facet $z \in Z$. Therefore, the distributional hypotheses (outlined in Section~\ref{subsec:distrhyp}) are intrinsically integrated in $\xi$ that is estimated from the reviews' text, conditioned on reviews' helpfulness scores. 

The estimated helpfulness score $\widehat{h}(u,i)$ of a review by user $u$ on item $i$ is a function $f$ of the following {\em consistency} and {\em latent} factors, parametrized by $\Psi$:
\vspace{-1.2em}

\begin{equation}
\label{eq:features}
 \widehat{h}(u,i) = f(\beta_u, \beta_i, |r-\overline{r_u}|, |r-\overline{r_i}|, |r-\overline{r_g}|, b_t, \xi; \Psi)
\end{equation}

Here, $f$ can be a polynomial, radial basis, or a simple linear function for combining the features. 
The objective is to estimate the parameters $\Psi$ (of dimension: $6 + E \times Z$) that reduces the {\em mean squared error} of the predicted helpfulness scores with the ground scores:
\vspace{-1em}

\begin{equation}
\label{eq:mse}
 \Psi^{*} = argmin_\Psi \frac{1}{|U|} \sum_{u,i \in U,I} (h(u,i) - \widehat{h}(u, i))^2 + \mu ||\Psi||^2_2
\end{equation}

\noindent where, we use $L_2$ regularization for the parameters to penalize complex models.
There are several ways to estimate the parameters like alternate least squares, gradient-descent, and Newton based approaches.

\vspace{-0.5em}
\subsection{Incorporating Latent Facets}
We use principles of {\em Latent Dirichlet Allocation} (LDA)~\cite{Blei2003LDA} to learn the latent facets associated to an item. Each review $d$ on an item is assumed to have a Multinomial distribution $\theta$ over facets $Z$ with a symmetric Dirichlet prior $\alpha$. Each facet $z$ has a Multinomial distribution $\phi_z$ over words drawn from a vocabulary $W$ with a symmetric Dirichlet prior $\delta$. 
Exact inference is not possible due to the intractable coupling between $\Theta$ and $\Phi$. Two popular ways for approximate inference are 
MCMC techniques like Collapsed Gibbs Sampling and Variational Inference.

\vspace{-0.5em}
\subsection{Incorporating Latent Expertise}

Expertise influences both the facet distribution $\Theta$, as users at different levels of expertise have different facet preferences, and the language model $\Phi$ as the writing style is also different for users at different levels of expertise. Therefore, we parametrize both of these distributions with user expertise similar to the prior work in~\cite{mukherjee2015jertm}, with some major modifications (discussed in the next section).

Consider $\Theta$ to be a tensor of dimension $E \times Z$, and $\Phi$ to be a tensor of dimension $E \times Z \times W$, where $\theta_{e,z}$ denotes the preference for facet $z \in Z$ for users at expertise level $e \in E$, and $\phi_{e,z,w}$ denotes the probability of the word $w \in W$ being used to describe the facet $z$ by users at expertise level $e$. 

Now, expertise changes as users evolve over time. However, the transition should be {\em smooth}. Users cannot abruptly jump from expertise level $1$ to $4$ without passing through expertise levels $2$ and $3$. Therefore, at each timepoint $t+1$ (of posting a review), we assume a user at expertise level $e_t \in E$ to stay at $e_t$, or move to $e_t + 1$ (i.e. expertise level is monotonically non-decreasing). This progression depends on how the writing style (captured by $\Phi$), and facet preferences (captured by $\Theta$) of the user is evolving {\em with respect to} other expert users in the community; as well as the rate of {\em activity} of the user 
--- that we use as a hyper-parameter for controlling the rate of progression. Let $\gamma_u$, the activity rate of user $u$, be defined as: $\gamma_u = \frac{D_
u}{D_u + D_{avg}}$,
where $D_u$ and $D_{avg}$ denote the number of posts written by $u$, and the average number of posts written by any user in the community, respectively.

Let $\Pi$ be a tensor of dimension $E \times E$ with hyper-parameters $\langle \gamma_u \rangle$ of dimension $U$, where $\pi_{e_i,e_j}$ denotes the probability of moving to expertise level $e_j$ from $e_i$ with the constraint $e_j \in \{e_i, e_i + 1\}$. However, not all users start at the same level of expertise, when they enter the community; some may enter already being an expert. The algorithm figures this out during the inference process. We assume all users to 
start at expertise level $1$ during parameter initialization.

During inference, we want to learn the parameters $\Psi, \xi, \Theta, \Phi, \Pi$ jointly for predicting review helpfulness. 

\subsection{Difference with Prior Works for Modeling Expertise} \label{subsec:diff}

Our generative process of user expertise is motivated by~\cite{mukherjee2015jertm,mcauleyWWW2013} with the following differences:\\
(i) The prior works learn {\em user-specific} preferences for personalized recommendation. However, we assume users at the same level of expertise to have similar facet preferences. Therefore, the facet distribution $\Theta$ is conditioned {\em only} on the user {\em expertise}, and not the user explicitly, unlike the prior works. This helps us to reduce the dimensionality of $\Theta$, and exploit the correspondence between $\Theta$ and $\xi$ to {\em tie} the parameters of the consistency and latent factor models together for tractable inference.\\
(ii) The prior work~\cite{mukherjee2015jertm} incorporates supervision, for predicting ratings, {\em only indirectly} via optimizing the Dirichlet hyper-parameters $\alpha$ of the Multinomial facet distribution $\Theta$ --- and cannot guarantee an increase in data log-likelihood over iterations. In contrast, we exploit (i) to learn the expertise-facet distribution $\Theta$ {\em directly} from the review helpfulness scores by minimizing the {\em mean squared error} during inference. This is also tricky as parameters of the distribution $\Theta$, for an unconstrained optimization, are not guaranteed to lie on the simplex --- for which we do certain transformations, discussed during inference. Therefore, parameters are {\em strongly} coupled in our model, not only reducing mean squared error, but also leading to a near smooth increase in data log-likelihood over iterations (refer Figure~\ref{fig:logL}). 

\subsection{Generative Process}
\vspace{-0.5em}

Consider a corpus $D=\{d_1,\ldots,d_D\}$ of reviews written by a set of users $U$ at timestamps $T$. For each review $d\in D$, we denote $u_d$ as its user, $t_d$ as the timestamp of the review.
Reviews are assumed to be ordered by timestamps, i.e., $t_{d_i}<t_{d_j}$ for $i<j$. 
 Each review $d \in D$ consists of a sequence of $N_d$ words denoted by $d=\{w_1,\ldots ,w_{N_d}\}$, where each word is drawn from a vocabulary $W$ with unique words indexed by $\{1 \dots W\}$, and $Z$ is the number of facets.
 
 Let $e_d \in \{1, 2, ..., E\}$ denote the expertise value of review $d$. Since each review $d$ is associated with a unique timestamp $t_d$ and unique user $u_d$, the expertise value of a review refers to the expertise of the user at the time of writing it. Following Markovian assumption, the user's expertise level transitions follow a distribution $\Pi$ with the Markovian assumption $e_{u_d} \sim \pi_{e_{u_{d-1}}}$ i.e. the expertise level of $u_d$ at time $t_d$ depends on her 
expertise level when writing the previous review at time $t_{d-1}$. 

Once expertise level $e_{d}$ of user $u_d$ for review $d$ is known, her facet preferences are given by $\theta_{e_d}$. Thereafter, the facet $z_{d,w}$ of each word $w$ in $d$ is drawn from a Multinomial ($\theta_{e_{d}}$). 
Now that the expertise level of a user, and her facets of interest are known, we can generate the language model $\Phi$ and individual words in the review ---  where the user draws a word from the Multinomial distribution $\phi_{e_{d}, z_{d,w}}$ with a symmetric Dirichlet prior $\delta$. Refer Figure~\ref{fig:plate} for the generative process.
%

\begin{figure}
	\centering
	\includegraphics[scale=0.35]{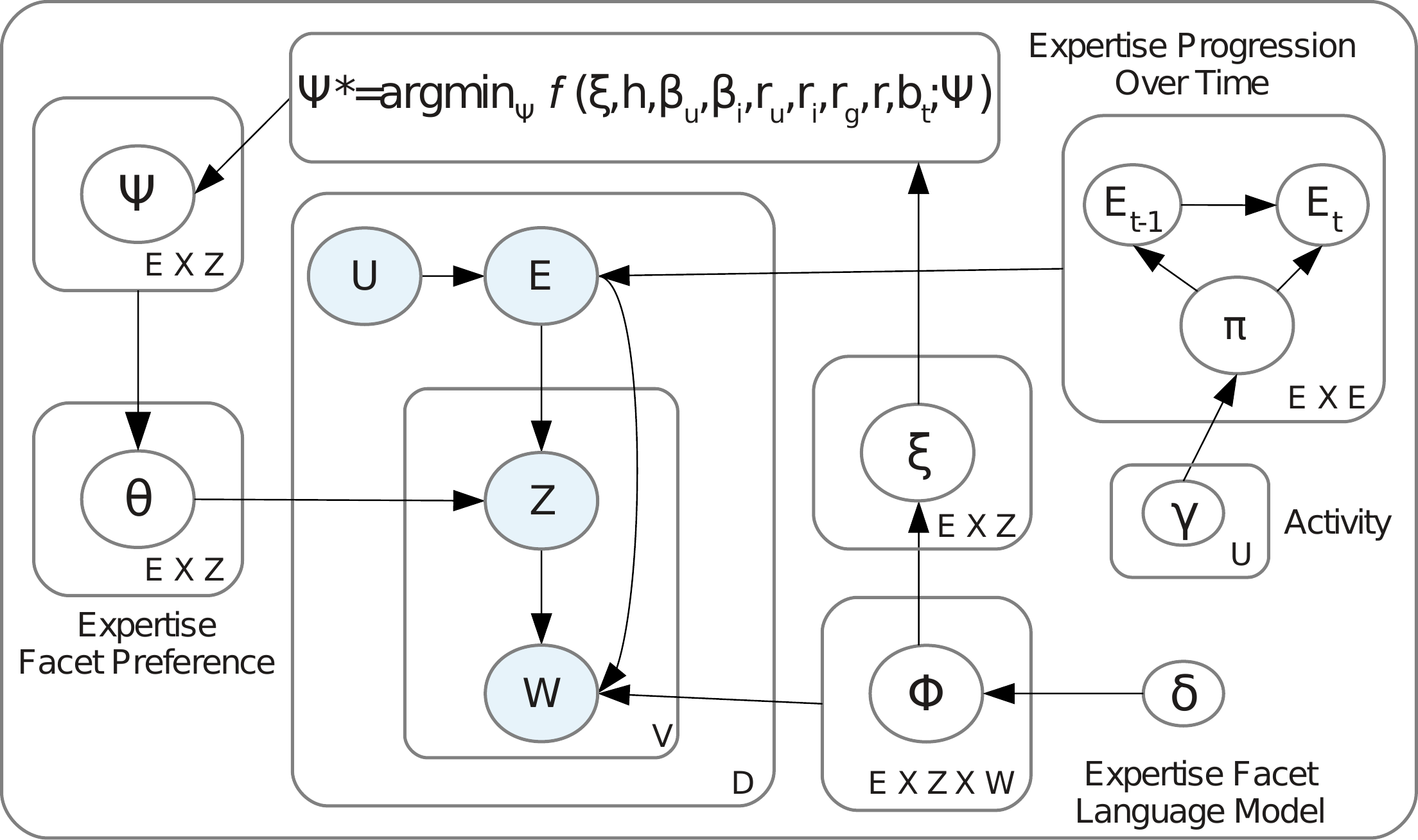}
	\caption{Generative process for helpful product reviews.}
	\label{fig:plate}
	\vspace{-1.5em}
\end{figure}

The joint probability distribution is given by:
\vspace{-1em}
{\small
\begin{multline}
\label{eq:joint}
 P(E,Z,W, \Theta, \Phi | U; \langle \gamma_u \rangle, \delta) \propto
 \prod_{u \in U} \prod_{d\in D_u}  P(\pi_{e_d}; \gamma_u) \boldsymbol{\cdot} P(e_d | \pi_{e_d}) \\ 
 \boldsymbol{\cdot} \bigg(\prod_{j=1}^{N_d} P(z_{d,j} | \theta_{e_d}) \cdot P(\phi_{e_d, z_{d,j}}; \delta) \cdot P(w_{d,j} | \phi_{e_d,z_{d,j}}) \bigg) 
\end{multline}
\vspace{-2em}
}

\subsection{Inference}

Given a corpus of reviews indexed by $\langle${\em userId, itemId, rating, reviewText, timepoint}$\rangle$, with corresponding helpfulness scores, our objective is to learn the parameters $\Psi$ that minimizes the mean squared error given by Equation~\ref{eq:mse}. 

In case $\xi$ was known, we could have directly plugged in its values (other features being {\em observed}) in Equation~\ref{eq:features} to learn a model (e.g., using regression) with parameters $\Psi$. However, the dimensions of $\xi$, corresponding to both facets and user expertise, are {\em latent} that need to be inferred from text. Now, the parameter $\psi_{e,z}$ corresponding to $\xi_{e,z}$ learned from Equation~\ref{eq:mse} depicts the importance of the facet $z$ for users at expertise level $e$ for predicting review helpfulness. We want to exploit this observation to infer the latent dimensions from text.

During the generative process of a review document, for a user at expertise level $e$, we want to draw her facet of interest $z$ with probability $\theta_{e,z} \propto \psi_{e,z}$. However, we cannot directly replace $\Theta$ with $\Psi$ due to the following reason. The traditional parametrization of a Multinomial distribution ($\Theta$ in this case) is via its mean parameters. Any unconstrained optimization will take the parameters out of the feasible set, i.e. they may not lie on the simplex. Hence, it is easier to work with the natural parameters instead. If we consider the unconstrained parameters $\langle \psi_{e,z} \rangle$ (learned from Equation~\ref{eq:mse}) 
to be the natural parameters of the Multinomial distribution $\Theta$, we need to transform the natural parameters to the mean parameters that lie on the simplex (i.e. $\sum_z \theta_{e,z} = 1$). In this work, we follow the principle similar to~\cite{mukherjee2016KDD} to do this transformation:
\vspace{-1em}

\begin{equation}
\label{eq:transform}
 \theta_{e,z} = \frac{exp(\psi_{e,z})}{\sum_z exp(\psi_{e,z})}
\end{equation}
\noindent 
where, $\psi_{e,z}$ corresponds to the learned parameter for $\xi_{e,z}$.


Exploiting conjugacy of the Multinomial and Dirichlet distributions, 
we can integrate out $\Phi$ from the joint distribution in Equation~\ref{eq:joint} to obtain the posterior distribution
$P(W|Z,E; \delta )$ given by:
\vspace{-1em}

{
\[\prod_{e=1}^E \prod_{z=1}^Z \frac{\Gamma(\sum_{w} \delta)\prod_{w} \Gamma(n(e, z, w)+ \delta)}{\prod_{w}{\Gamma(\delta)\Gamma(\sum_{w} n(e, z, w) + \sum_w \delta)}}\]
} 

\noindent where, $\Gamma$ denotes the Gamma function, and $n(e,z,w)$ is the number of times the word $w$ is used for facet $z$ by users at expertise level $e$.

We use Collapsed Gibbs Sampling~\cite{Griffiths02gibbssampling}, as in standard LDA,
to estimate the conditional distribution for each of the latent facets $z_{d,j}$,
which is computed over the current assignment for all other hidden
variables, after integrating out $\Phi$. In the following equation,
 $n(e,z,.)$  indicates the summation of the counts over all possible $w\in W$.  
The subscript $-j$ denotes the value
of a variable excluding the data at the $j^{th}$ position.

The posterior distribution
$P(Z| \Phi, W, E)$ of the latent variable $Z$ is given by:

\begin{equation}
\begin{aligned}
\label{eq:facets}
 & P(z_{d,j} = k | z_{d,-j}, \Phi, w_{d,j}=w, e_d=e, d) \\
 & \propto \theta_{e,k} \boldsymbol{\cdot} \frac{n(e, k, w) + \delta}{n(e,k, .) + W \cdot \delta} \\
 & = \frac{exp(\psi_{e,k})}{\sum_z exp(\psi_{e,z})} \boldsymbol{\cdot} \frac{n(e, k, w) + \delta}{n(e,k, .) + W \cdot \delta} \\
\end{aligned} 
\end{equation}

Similar to the above process, we use Collapsed Gibbs Sampling \cite{Griffiths02gibbssampling} also to sample the expertise levels, keeping all facet assignments $Z$ fixed. Let $n(e_{i-1}, e_i)$ denote the number of transitions from expertise level $e_{i-1}$ to $e_i$ over {all} users in the community, with the Markovian constraint $e_i \in \{e_{i-1}, e_{i-1}+1\}$. 

\begin{equation}
\label{eq:transition}
P(e_i|e_{i - 1}, e_{-i}, u; \gamma_u) = \frac{n(e_{i - 1}, e_i) + I(e_{i - 1} = e_i) + \gamma_u}{n(e_{i-1},.) + I(e_{i - 1} = e_i) + E \boldsymbol{\cdot}\gamma_u} 
\end{equation}
\noindent where $I(.)$ is an indicator function taking the value $1$ when the argument is true (a self-transition, in this case, where the user has the same expertise level over subsequent reviews), and $0$ otherwise. The subscript $- i$ denotes the value of a variable excluding the data at the $i^{th}$ position. Note that the transition function is similar to prior works in Hidden Markov Model -- Latent Dirichlet Allocation (HMM-LDA) based models~\cite{rosenzviUAI2004},~\cite{mukherjeeSDM2014}.

The conditional distribution for the expertise level transition is given by:

\begin{equation}
\label{eq:expertise}
 P(E|U,Z,W; \langle \gamma_u \rangle) \propto P (E|U; \langle \gamma_u \rangle) \boldsymbol{\cdot} P(Z|E) \boldsymbol{\cdot} P(W|Z, E)\\
\end{equation}
Using Equations~\ref{eq:facets},~\ref{eq:transition},~\ref{eq:expertise}, we obtain the conditional distribution for updating latent variables $E$ as:
\begin{equation}
\small
\begin{aligned}
\label{eq:exp-gibbs}
\hspace{-1em}
& P(e_{u_d} = e_i |e_{u_{d-1}}=e_{i-1}, u_d = u, \{z_{i,j}=z_j \}, \{w_{i,j}=w_j\}, e_{-i}) \\
& \propto \frac{n(e_{i - 1}, e_i) + I(e_{i - 1} = e_i) + \gamma_u}{n(e_{i-1}, .) + I(e_{i - 1} = e_i) + E \boldsymbol{\cdot}\gamma_u} \\
& \boldsymbol{\cdot} \bigg( \prod_j  \frac{exp(\psi_{e_i,z_j})}{\sum_z exp(\psi_{e_i,z})} \boldsymbol{\cdot} \frac{n(e_i, z_j, w_j) + \delta}{n(e_i,z_j, .) + W \cdot \delta} \bigg) \\
\end{aligned}
\end{equation}

Consider a document $d$ containing a sequence of words $\{w_j\}$ with corresponding facets $\{z_j\}$. The first factor models the probability of the user $u_d$ reaching expertise level $e_{u_d}$ for document $d$; whereas the second and third factor models the probability of the facets $\{z_j\}$ being chosen at the expertise level $e_{u_d}$, and the probability of observing the words $\{w_j\}$ with the facets $\{z_j\}$ and expertise level $e_{u_d}$, respectively. Following the Markovian assumption, we only consider the expertise levels $e_{u_d}$ and $e_{u_d} + 1$ for sampling, and select the one with the highest conditional probability.

Samples obtained from Gibbs sampling are
used to approximate the expertise-facet-word distribution
$\Phi$:

\begin{equation}
\label{eq:phi}
 \phi_{e,z,w} = \frac{n(e, z, w) + \delta}{n(e,z, .) + W \cdot \delta}
\end{equation}

Once the generative process for a review $d$ with words $\{w_j\}$ is over, we can estimate $\xi$ from $\Phi$ as the proportion of the $z^{th}$ facet in the document written at expertise level $e$ as:

\begin{equation}
\label{eq:xi}
\xi_{e,z} \propto \sum_{j=1}^{N_d} \phi_{e, z, w_j}
\end{equation}

In summary, $\xi$, $\Phi$, and $\Theta$ are linked via $\Psi$:
\squishlisttwo
\item[i)] $\Psi$ generates $\Theta$ via Equation~\ref{eq:transform}.
\item[ii)] $\Theta$ and $\Phi$ are coupled in Equations~\ref{eq:joint},~\ref{eq:facets}.
\item[iii)] $\Phi$ generates $\xi$ using Equation~\ref{eq:xi}.
\item[iv)] $\Psi$ is learned via regression (with $\xi$ as latent features) using Equations~\ref{eq:features},~\ref{eq:mse}, so as to minimize the mean squared error for predicting review helpfulness. 
\squishend

\noindent {\bf Overall Processing Scheme:} Exploiting results from the above discussions, the overall inference is an iterative stochastic optimization process consisting of the following steps:
\squishlisttwo
\item[i)] Sort all reviews by timestamps, and estimate $E$ using Equation~\ref{eq:exp-gibbs}, by Gibbs sampling. During this process, consider all facet assignments $Z$ and $\Psi$, from the earlier iteration fixed.
\item[ii)] Estimate facets $Z$ using Equation~\ref{eq:facets}, by Gibbs sampling, keeping the expertise levels $E$ and $\Psi$, from the earlier iteration fixed.
\item[iii)] Estimate $\xi$ using Equations~\ref{eq:phi} and~\ref{eq:xi}.
\item[iv)] Learn $\Psi$ from $\xi$ and other consistency factors using Equations~\ref{eq:features},~\ref{eq:mse}, by regression.
\item[v)] Estimate $\Theta$ from $\Psi$ using Equation~\ref{eq:transform}.
\squishend

\noindent{\bf Regression:} For regression, we use the fast and scalable Support Vector Regression implementation from LibLinear ({\tt \small www.csie.ntu.edu.tw/~cjlin/liblinear}) that uses Trust Region Newton method for learning the parameters $\Psi$.

\noindent{\bf Test:} Given a test review with $\langle${\em user=$u$, item=$i$, words=$\{w_j \}$, rating=$r$, timestamp=$t$}$\rangle$, we find its helpfulness score by plugging in the consistency features, and latent factors in Equation~\ref{eq:features} with the parameters $\langle \Psi, \beta_u, \beta_i, \overline{r_u}, \overline{r_i}, \overline{r_g} \rangle$ having been learned from the training data. $\xi$ is computed over the words $\{w_j\}$ using Equation~\ref{eq:xi} --- with the distribution $\Phi$ having been learned during training.


\section{Experiments}
\label{sec:experiments}

\noindent {\bf Setup:} We perform experiments with data from {\em Amazon} in {\em five} different domains:
(i) movies, (ii) music, (iii) food, (iv) books, and (v) electronics. The statistics of the dataset ({\tt \small snap.stanford.edu/data/}) is given in Table~\ref{tab:statistics}. 
In total, we have $29$ million reviews from $5.6$ million users on $1.8$ million items from all of the five domains combined. We extract the following quintuple for our model $\langle${\em userId, itemId, timestamp, rating, review, helpfulnesVotes}$\rangle$ from each domain. 
During {\em training}, for movies, books, music, and electronics, we consider only those reviews for which at least $y \geq 20$ users have voted about their helpfulness (including for, and against) to have a robust dataset (similar to the setting in~\cite{icdm2008, O'Mahony:2009:LRH:1639714.1639774}) for learning. Since the food dataset has less number of reviews, we lowered this threshold to {\em five}. 
For {\em test}, we used the $3$ most recent reviews of each user as withheld test data (similar to the setting in~\cite{mukherjee2015jertm, mcauleyWWW2013}), that received atleast {\em five} votes (including for, and against). {\em The same data is used for all the models for comparison}.
We group {\em long-tail} users with less than $10$ reviews in {\em training} data into a background model, treated as a single user, to avoid modeling from sparse observations. We do not ignore any user. During the {\em test} phase for a ``long-tail'' user, we take her parameters from the background model. We set the number of facets as $Z=50$, and number expertise levels as $E=5$, for all the datasets. 

\begin{table}[t]
	\centering
	\resizebox{\linewidth}{!}{%
	\begin{tabular}{lrrrrr}
		\toprule
		\bf{Factors} & \bf{Books} & \bf{Music} & \bf{Movie} & \bf{Electronics} & \bf{Food}\\
		\midrule
		\#Users & 2,588,991 & 1,134,684 & 889,176 & 811,034 & 256,059\\
		\#Items & 929,264 & 556,814 & 253,059 & 82,067 & 74,258\\
		\#Reviews & 12,886,488 & 6,396,350 & 7,911,684 & 1,241,778 & 568,454\\
		\midrule
		$\frac{\textbf{\#Reviews}}{\textbf{\#Users}}$ & 4.98 & 5.64 & 8.89 & 1.53 & 2.22 \\\midrule
		$\frac{\textbf{\#Reviews}}{\textbf{\#Items}}$ & 13.86 & 11.48 &	31.26 &	15.13 &	7.65 \\\midrule
		$\frac{\textbf{\#Votes}}{\textbf{\#Reviews}}$ & 9.71 & 5.95 & 7.90 & 8.91 & 4.24\\
		\bottomrule
	\end{tabular}
	}
	\vspace{-0.5em}
	\caption{\small Dataset statistics. Votes indicate the total number of helpfulness votes (both, for and against) cast for a review. Total number of users $=5,679,944$, items $=1,895,462$, and reviews $=29,004,754$.}
	\label{tab:statistics}
	\vspace{-2em}
\end{table}

\noindent {\bf Tasks and Evaluation Measures:} 
We use all the models for the following tasks:\\
1) {\bf Prediction:} Here the objective is to predict the helpfulness score of a review as $x/y$, where $x$ is the number of users who voted the review as helpful out of $y$ number of users. As evaluation measures, we report:\\ (i) {\em Mean squared error} of the predicted scores with the ground helpfulness scores (using Equation~\ref{eq:mse}), and (ii) {\em Squared correlation coefficient ($R^2$)} that gives an indication of the goodness of fit of a model, i.e., how well the regression function approximates the real data points, with $R^2 =1$ indicating a perfect fit.\\
2) {\bf Ranking:} A more suitable way of evaluation is to compare the ranking of reviews from different models sorted on their (predicted) helpfulness scores --- where the reviews at the top of the rank list should be more helpful than the ones below --- and compute {\em rank correlation} with the gold/reference rank list (sorted by ground-truth helpfulness scores $x/y$) using the following measures: (i) {\em Spearman correlation ($\rho$)} that assesses how well the relationship between two variables can be described using a {\em monotonic} function, 
and (ii) {\em Kendall-Tau correlation ($\tau$)} that measures the number of concordant and discordant pairs, to find whether the ranks of two elements agree or not based on their scores, out of the total number of combinations possible. 

\subsection{Quantitative Comparison}

\begin{table*}[t]
	\centering
	\scriptsize
	\resizebox{\linewidth}{!}{%
		\begin{tabular}{p{3.5cm}p{0.8cm}p{0.7cm}p{0.7cm}p{0.6cm}p{1.3cm}|p{0.8cm}p{0.7cm}p{0.7cm}p{0.6cm}p{1.3cm}}
			\toprule
			\multirow{2}{*}{\bf{Models}} & \multicolumn{5}{c|}{\textbf{Mean Squared Error (MSE)}} &  \multicolumn{5}{c}{\textbf{Squared Correlation Coefficient ($R^2$)}} \\ \cmidrule{2-11}
			& \bf{Movies} & \bf{Music} & \bf{Books} & \bf{Food}  & \bf{Electr.} & \bf{Movies} & \bf{Music} & \bf{Books} & \bf{Food}  & \bf{Electr.}\\
			\midrule
			Our model & {\bf 0.058}  & {\bf 0.059} & {\bf 0.055} & {\bf 0.053}  & {\bf 0.050} & {\bf 0.438} & {\bf 0.405} & {\bf 0.397} & {\bf 0.345}  & {\bf 0.197}\\
			{\bf a)} P.O'Mahony et al.~\cite{O'Mahony:2009:LRH:1639714.1639774} & 0.067 & 0.069 &  0.069 & 0.060 & 0.064 & 0.325 & 0.295 & 0.249 & 0.312 & 0.134\\
			{\bf b)} Lu et al.~\cite{Lu:2010:ESC:1772690.1772761} & 0.093 & 0.087 & 0.077 & 0.072 & 0.071 & 0.111 & 0.128 & 0.139 & 0.134 & 0.056 \\
			{\bf c)} Kim et al.~\cite{Kim:2006:AAR:1610075.1610135} & 0.107 & 0.125 & 0.094 & 0.073 & 0.161 & 0.211 & 0.025 & 0.211 & 0.309 & 0.065 \\
			{\bf d)} Liu et al.~\cite{icdm2008} & 0.091 & 0.091 & 0.082 & 0.075 & 0.063 & 0.076 & 0.053 & 0.076 & 0.039 & 0.043 \\
			\bottomrule
		\end{tabular}
	}
	\vspace{-1em}
	\caption{\small {\em Prediction Task:} Performance comparison of our model versus baselines. Our {improvements} over the baselines are statistically significant at {\em p-value} $<2.2e-16$ using {\em paired sample t-test}.}
	\label{fig:MSE}
	\vspace{1em}
	\resizebox{\linewidth}{!}{%
		\begin{tabular}{p{3.5cm}p{0.8cm}p{0.7cm}p{0.7cm}p{0.6cm}p{1.3cm}|p{0.8cm}p{0.7cm}p{0.7cm}p{0.6cm}p{1.3cm}}
			\toprule
			\multirow{2}{*}{\bf{Models}} & \multicolumn{5}{c|}{\textbf{Spearman ($\rho$)}} &  \multicolumn{5}{c}{\textbf{Kendall-Tau ($\tau$)}} \\ \cmidrule{2-11}
			& \bf{Movies} & \bf{Music} & \bf{Books} & \bf{Food}  & \bf{Electr.} & \bf{Movies} & \bf{Music} & \bf{Books} & \bf{Food} & \bf{Electr.}\\
			\midrule
			Our model & {\bf 0.657} & {\bf 0.610} &  {\bf 0.603} & 0.533  & {\bf 0.394} & {\bf 0.475} & {\bf 0.440} & {\bf 0.435} & 0.387  & {\bf 0.280}\\
			{\bf a)} P.O'Mahony et al.~\cite{O'Mahony:2009:LRH:1639714.1639774} & 0.591 & 0.554  & 0.496 & 0.541 & 0.340 & 0.414 & 0.390 & 0.347 & 0.398 & 0.237\\
			{\bf b)} Lu et al.~\cite{Lu:2010:ESC:1772690.1772761} & 0.330  & 0.349 & 0.334 & 0.367 & 0.205 & 0.224 & 0.242 & 0.230 & 0.259 & 0.144 \\
			{\bf c)} Kim et al.~\cite{Kim:2006:AAR:1610075.1610135} & 0.489 & 0.166 & 0.474 & \textbf{0.551} & 0.261 & 0.342 & 0.114 & 0.334 & \textbf{0.414}  & 0.184 \\
			{\bf d)} Liu et al.~\cite{icdm2008} & 0.268 & 0.232 & 0.258 & 0.199 & 0.159 & 0.183 & 0.161 & 0.178 & 0.141 & 0.112 \\
			\bottomrule
		\end{tabular}
	}
	\vspace{-1em}
	\caption{\small {\em Ranking Task:} Correlation comparison between the ranking of reviews and gold rank list --- our model versus baselines. Our {\em improvements} over the baselines are statistically significant at {\em p-value} $<2.2e-16$ using {\em paired sample t-test}.}
	\label{fig:rank}
	\vspace{1em}
	\begin{tabular}{p{17cm}}
	\toprule
	{\bf Top words used by experts in {\em most} helpful reviews.}\\
	\midrule
	  {\bf Music:} album, lyrics, recommend, soundtrack, touch, songwriting, features, rare, musical, ears, lyrical, enjoy, absolutely, musically, individual, bland, soothing, released, inspiration, share, mainstream, deeper, flawless, wonderfully, eclectic, heavily, critics, presence
	  \\
	  {\bf Books:} serious, complex, claims, content, illustrations, picture, genre, beautifully, literary, witty, critics, complicated, argument, premise, scholarship, talented, divine, twists, exceptional, obsession, commentary, landscape, exposes, influenced, accomplished, oriented
	  \\
	  {\bf Movies:} scene, recommend, screenplay, business, depth, justice, humanity, packaging, perfection, flicks, sequels, propaganda, anamorphic, cliche\&acute, pretentious, goofy, ancient, marvelous, perspective, outrageous, intensity, mildly, immensely, bland, subplots, anticipation
	  \\
	  {\bf Electronics:} adapter, wireless, computer, sounds, camera, range, drives, mounted, photos, shots, packaging, antenna, ease, careful, broken, cards, distortion, stick, media, application, worthless, clarity, technical, memory, steady, dock, items, cord, systems, amps, skin, watt
	  \\
	  {\bf Food:} expensive, machine, months, clean, chips, texture, spicy, odor, inside, processed, robust, packs, weather, sticking, alot, press, poured, swallow, reasonably, portions, beware, fragrance, basket, volume, sweetness, terribly, caused, scratching, serves, sensation, sipping, smelled
	  \\\toprule
	  	{\bf Top words used by amateurs in {\em least} helpful reviews.}\\
	\midrule
	  {\bf Music:} will, good, favorite, cool, great, genius, earlier, notes, attention, place, putting, superb, style, room, beauty, realize, brought, passionate, difference, god, fresh, save, musical, grooves, consists, tapes, depressing, interview, short, rock, appeared, learn, brothers
	  \\
	  {\bf Books:} will, book, time, religious, liberal, material, interest, utterly, moves, movie, consistent, false, committed, question, turn, coverage, decade, novel, understood, worst, leader, history, kind, energy, fit, dropped, current, doubt, fan, books, building, travel, sudden, fails
	  \\
	  {\bf Movies:} movie, hour, gay, dont, close, previous, features, type, months, meaning, wait, boring, absolutely, truth, generation, going, fighting, runs, fantastic, kids, quiet, kill, lost, angles, previews, crafted, teens, help, believes, brilliance, touches, sea, hardcore, continue, album
	  \\
	  {\bf Electronics:} order, attach, replaced, write, impressed, install, learn, tool, offered, details, turns, snap, price, digital, well, buds, fit, problems, photos, hear, shoot, surprisingly, continue, house, card, sports, writing, include, adequate, nice, programming, protected, mistake
	  \\
	  {\bf Food:} night, going, haven, sour, fat, avoid, sugar, coffee, store, bodied, graham, variety, salsa, reasons, favorite, delicate, purpose, brands, worst, litter, funny, partially, sesame, handle, excited, close, awful, happily, fully, fits, effects, virgin, salt, returned, powdery, meals, great
	  \\\bottomrule
	\end{tabular}
		\vspace{-1.2em}
	\caption{\small Snapshot of latent word clusters as used by experts and amateurs for most and least helpful reviews in $5$ domains.} \label{tab:examples}
	\vspace{-2em}
\end{table*}


\noindent {\bf Baselines:} We consider the following baselines to compare our work:\\
{\em (a) P.O'Mahony et al. (RecSys, 2009)~\cite{O'Mahony:2009:LRH:1639714.1639774}} use several rating based features as proxy for reviewer reputation and sentiment; review length and letter cases for content; and review count statistics for social features.
\\ 
{\em (b) Lu et al. (WWW, 2010)~\cite{Lu:2010:ESC:1772690.1772761}} use syntactic features (part-of-speech tags of words), sentiment (using a lexicon to find word polarities), review length and reviewer rating statistics to predict the quality of a review. 
\\
{\em (c) Kim et al. (EMNLP, 2006)~\cite{Kim:2006:AAR:1610075.1610135}} use structural (review length statistics), lexical (tf-idf), syntactic (part-of-speech tags), semantic (explicit product features, and sentiment of words), and meta-data related features to rank the reviews based on their helpfulness. 
\\
{\em (d) Liu et al. (ICDM 2008)~\cite{icdm2008}} predict the helpfulness of reviews on IMDB based on: \textit{reviewer expertise}, \textit{syntactic features}, and \textit{timeliness} of a review. The authors use reviewer preferences for explicit facets (pre-defined genres of movies in IMDB) as {\em proxy} for their expertise, part-of-speech tags of words for syntactic features, and review publication dates for ``timeliness'' of reviews. This baseline is the closest to our work as we attempt to model similar factors. However, we model reviewer expertise {\em explicitly}, and the facets as {\em latent} --- therefore not relying on any additional item meta-data (like, genres) or proxies for user authority.

For all of the above baselines, we use all the features from their works that are supported by our dataset --- for instance, we could not use the social network, and explicit product meta-data absent in our dataset --- for a fair comparison. 
Table~\ref{fig:MSE} shows the comparison of the \emph{Mean Squared Error (MSE)} and \textit{Squared Correlation Coefficient ($R^2$)} for review helpfulness predictions, as  generated by our model with the four baselines. Our model consistently outperforms all baselines in reducing the MSE.
Table~\ref{fig:rank} shows the comparison of the \textit{Spearman ($\rho$)} and \textit{Kendall-Tau ($\tau$)} correlation between the rank list of helpful user reviews, as generated by all the models, and the gold rank list.

The most competitive baseline for our model is~\cite{icdm2008}. Due to a {\em high overlap in consistency features} of our model with this baseline, our performance improvement can be attributed to the {\em incorporation of latent factors} in our model. We perform {\em paired sample t-tests}, and find that our performance improvement over all the baselines is statistically significant at {\em p-value} $<2e-16$. We perform the best for the domains {\em movies, music,} and {\em books} with large number of reviews, and relatively worse in the domains {\em food, electronics} due to data sparsity for which user maturity could not be captured well. 


\begin{figure*}[t]
	\centering
	\includegraphics[scale=0.7]{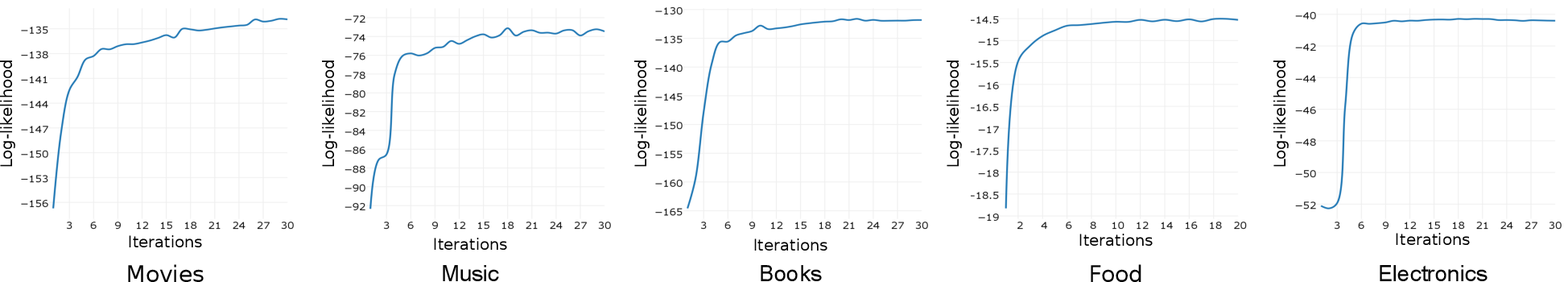}
		\vspace{-.5em}
	\caption{\small Increase in log-likelihood (scaled by $10e+07$) of the data {\em per-iteration} in the five domains.}
	\label{fig:logL}
	\vspace{-1.3em}
\end{figure*}
\begin{figure*}
\centering
	\subfloat[\small Our model: Facet preference divergence with expertise learned from review helpfulness.]{	\includegraphics[scale=0.7]{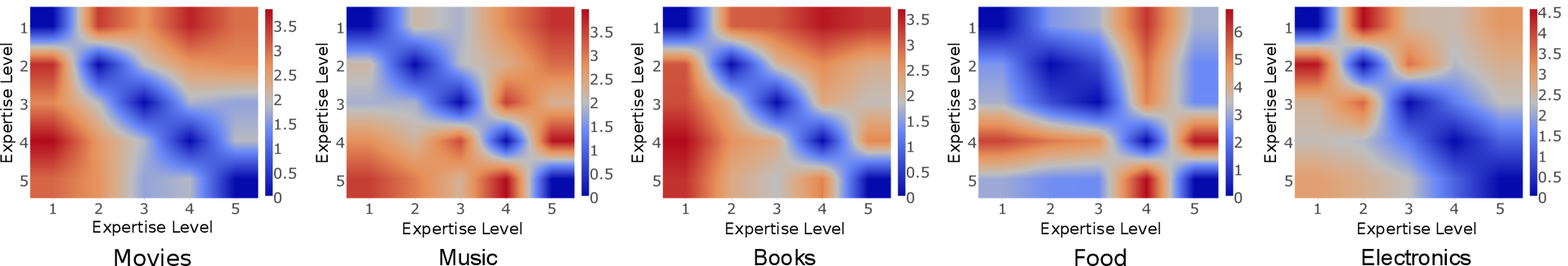}\label{fig:heatMap_facet}}
	\vspace{-0.3cm}
	\subfloat[\small Our model: Language model divergence with expertise learned from review helpfulness.]{	\includegraphics[scale=0.7]{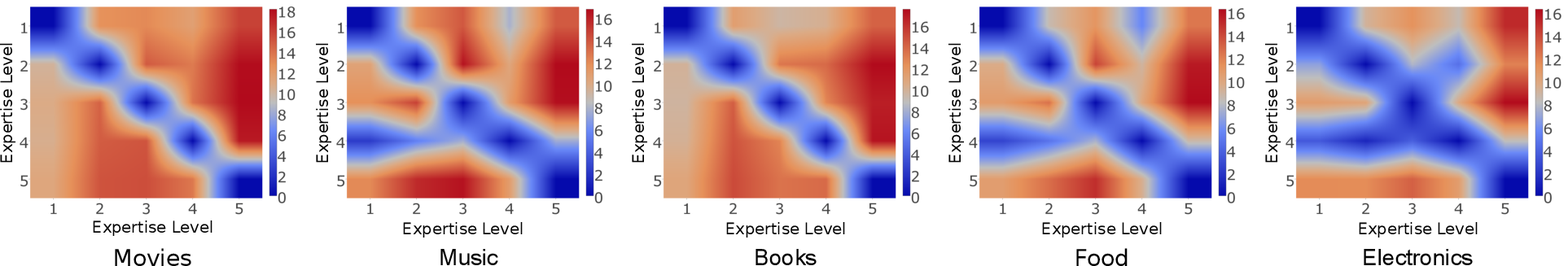}\label{fig:heatMap_lang}}
	\vspace{-.5em}
	\caption{Facet preference and language model KL divergence with expertise.}
	\vspace{-1.5em}
\end{figure*}

\subsection{Qualitative Comparison\\}

{\noindent \bf Log-likelihood of data and convergence:} The inference of our model is quite involved with the coupling between several variables, and the alternate stochastic optimization process. Figure~\ref{fig:logL} shows the increase in the data log-likelihood of our model per-iteration for different domains. We observe that the model is stable, and achieves a near smooth increase in the data log-likelihood per-iteration. It also converges quite fast between $20-30$ iterations depending on the complexity of the dataset. For {\em electronics} the convergence is quite rapid as the data is quite sparse, and the model does not find sufficient evidence for categorizing users to different expertise levels; this behavior is reflected in all the experiments involving the {\em electronics} dataset.

{\noindent \bf Language model and facet preference divergence:} 
Figures~\ref{fig:heatMap_facet} and~\ref{fig:heatMap_lang} show the heatmaps of the Kullback-Leibler (KL) divergence for facet preferences and language models of users at different expertise levels, as computed by our model conditioned on review {\em helpfulness} --- given by $D_{KL}(\theta_{e_i}||\theta_{e_j})$ and $D_{KL}(\phi_{e_i}||\phi_{e_j})$ respectively, where $\Theta$ and $\Phi$ are given by Equations~\ref{eq:transform} and~\ref{eq:phi}, respectively. 

The main observation is that the {\em KL} divergence is higher --- the larger the difference is between the expertise levels of two users. This confirms our hypothesis that expert users have a distinctive writing style and facet preferences different than that of amateurs --- captured by the joint interactions between review helpfulness, reviewer expertise, facet preferences, and writing style. We also note that the increase in divergence with the increase in gap between expertise levels is not smooth for {\em food} and {\em electronics} due to sparsity of {\em per-user} data.

{\noindent \bf Interpretable explanation by salient words used by experts for helpful reviews:} Table~\ref{tab:examples} shows a snapshot of the latent word clusters, as used by experts and amateurs, for helpful reviews and otherwise, as generated by our model. 

We observe that the most helpful reviews pertaining to {\em music} talk about its essence and style; for {\em books} they describe the theme and writing style; for {\em movies} they write about screenplay and storytelling; for {\em food} reviews these are mostly concerned about hygiene and allergens; for {\em electronics} they discuss about specific product features. {\em Note} that prior works~\cite{icdm2008,Kim:2006:AAR:1610075.1610135} used {\em explicit} product features, that we were able to automatically discover as latent features from reviews. The least helpful reviews mostly describe some generic concepts, praise or criticize an item without going in depth about the facets, and are generally quite superficial in nature.

\section{Conclusion}\label{sec:conc}

We proposed an approach to predict useful product reviews by exploiting the {\em joint interaction} between user expertise, writing style, timeliness, and review consistency using Hidden Markov Model -- Latent Dirichlet Allocation. Unlike prior works exploiting a variety of syntactic and domain-specific features, our model uses {\em only} the information of a user reviewing an item at an explicit timepoint for this task --- making our approach generalizable across all communities and domains. Additionally, we provide {\em interpretable explanation} as to why a review is helpful, in terms of salient words from latent word clusters --- that are used by experts to describe important facets of the item in the review.
Our experiments on real-world datasets from Amazon like books, movies, music, food, and electronics demonstrate effectiveness of our approach over state-of-the-art baselines.

\section*{Acknowledgment}
This research was partly supported 
by ERC Synergy Grant 610150 (imPACT). 

\bibliographystyle{abbrv}
\bibliography{sdm16}

\end{document}